%% file: main.tex
\documentclass[10pt,twocolumn,letterpaper]{article}

\usepackage{cvpr}      

\def\cvprPaperID{5148} 
\def\confName{CVPR}
\def\confYear{2022}
\usepackage{times}
\usepackage{epsfig}
\usepackage{graphicx}
\usepackage{amsmath}
\usepackage{amssymb}
\usepackage{url}            
\usepackage{booktabs}       
\usepackage{amsfonts}       
\usepackage{nicefrac}       
\usepackage{microtype}      
\usepackage{times}
\usepackage{epsfig}
\usepackage{graphicx}
\usepackage{amsmath}
\usepackage{amssymb}
\usepackage{caption}
\usepackage{subcaption}
\usepackage{algorithm}
\usepackage{algorithmic}
\usepackage{wrapfig}
\usepackage{lipsum}
\usepackage{bbm}
\usepackage[pagebackref,breaklinks,colorlinks]{hyperref}

\usepackage[capitalize]{cleveref}
\crefname{section}{Sec.}{Secs.}
\Crefname{section}{Section}{Sections}
\Crefname{table}{Table}{Tables}
\crefname{table}{Tab.}{Tabs.}

\begin{document}

\title{Canonical Voting: Towards Robust Oriented Bounding Box Detection \\in 3D Scenes}

\author{Yang You, Zelin Ye, Yujing Lou, Chengkun Li, Yong-Lu Li, Lizhuang Ma, Weiming Wang\thanks{Cewu Lu and Weiming Wang are the corresponding authors. Cewu Lu is member of Qing Yuan Research Institute and MoE Key Lab of Artificial Intelligence, AI Institute, Shanghai Jiao Tong University, China and Shanghai Qi Zhi institute.},\, Cewu Lu\footnotemark[1]  \\ 
Shanghai Jiao Tong University, China\\ 
\{qq456cvb, h\_e\_r\_o, louyujing, sjtulck, yonglu\_li, ma-lz, wangweiming, lucewu\}@sjtu.edu.cn 
}

\maketitle

\begin{abstract}
3D object detection has attracted much attention thanks to the advances in sensors and deep learning methods for point clouds. Current state-of-the-art methods like VoteNet regress direct offset towards object centers and box orientations with an additional Multi-Layer-Perceptron network. Both their offset and orientation predictions are not accurate due to the fundamental difficulty in rotation classification. In the work, we disentangle the direct offset into Local Canonical Coordinates (LCC), box scales and box orientations. Only LCC and box scales are regressed, while box orientations are generated by a canonical voting scheme. Finally, an LCC-aware back-projection checking algorithm iteratively cuts out bounding boxes from the generated vote maps, with the elimination of false positives. Our model achieves state-of-the-art performance on three standard real-world benchmarks: ScanNet, SceneNN and SUN RGB-D. Our code is available on \href{https://github.com/qq456cvb/CanonicalVoting}{https://github.com/qq456cvb/CanonicalVoting}.
\end{abstract}

\section{Introduction}
With the use of depth cameras and Lidar sensors, 3D object detection is becoming more and more important for real-world scene understanding. Recently, with advances in deep networks for point clouds, several methods~\cite{qi2019deep,shi2019pointrcnn,song2016deep,xu2018pointfusion} have shown state-of-the-art 3D detection results. Among them, the recently proposed
VoteNet~\cite{qi2019deep} showed remarkable improvement over previous methods on 3D oriented bounding box detection (rotation around gravity axis).

\begin{figure}[ht]
    \centering
    \includegraphics[width=\linewidth]{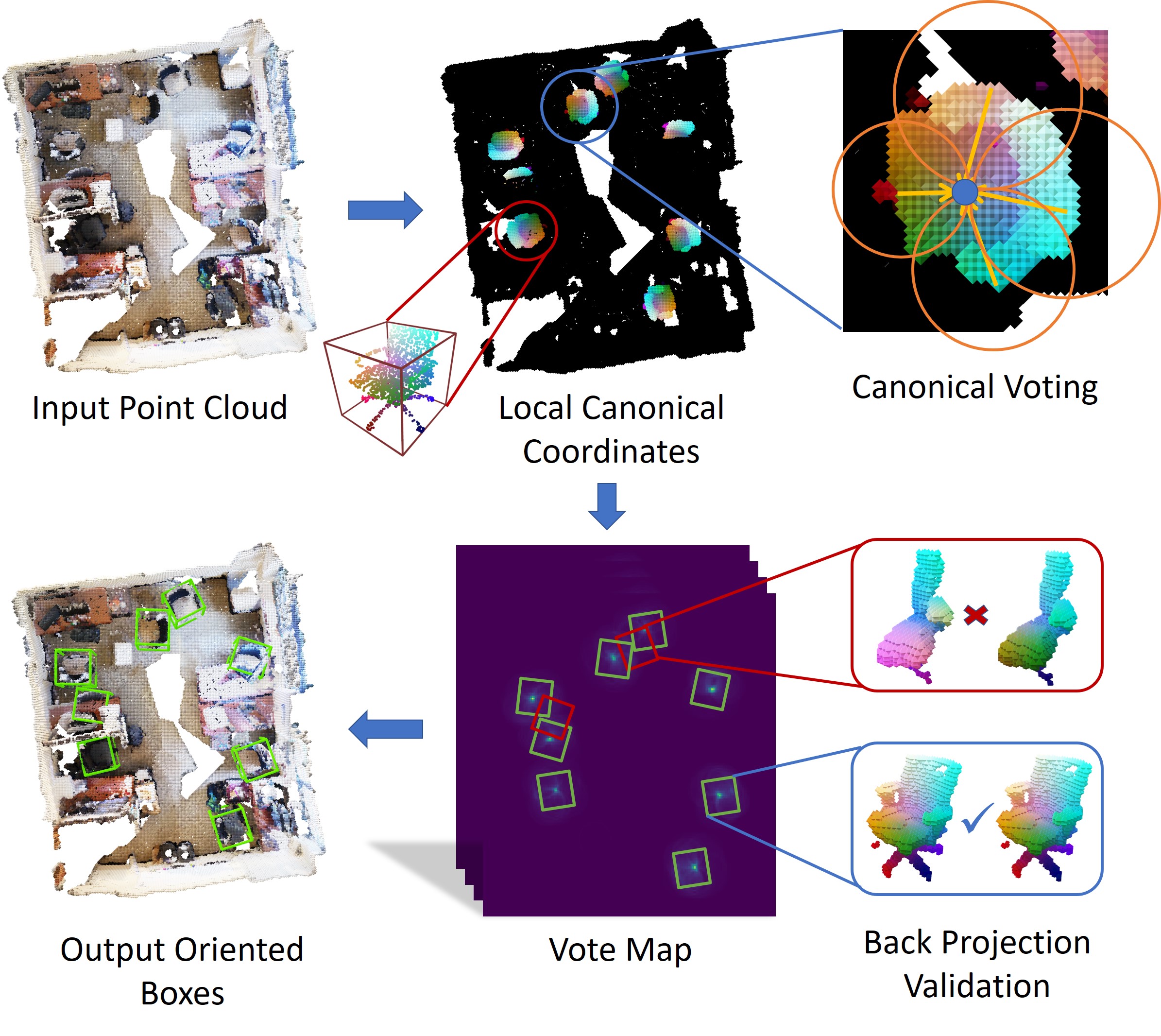}
    \caption{\textbf{We present a method that regresses Local Canonical Coordinates disentangled from orientations.} We leverage a canonical voting module to find possible orientations and object centers. Back projection validation is utilized to further eliminate false positives.}
    \label{fig:intro}
\end{figure}


VoteNet passes the input point cloud through a backbone network and then samples a set of seed points, which generate center votes. These votes are offset, targeted to reach object centers. After that, vote clusters are aggregated through a learned module to generate box orientations and scales. VoteNet resembles with traditional Hough voting in that bounding box centers are proposed in those peaks of votes. 
However, we find that neither box orientations nor offsets towards object centers are accurately predicted for most points, even with modern 3D sparse convolution techniques~\cite{choy20194d}. The absolute error of point-wise predicted offsets and box orientations is not even better than random guess in many cases, as shown in Table~\ref{tab:intro}. Though VoteNet proposes a two-stage pipeline and refines the center and orientation in the later stage, the accumulated error cannot be eliminated.

\begin{table}[h]
    \begin{center}
    \resizebox{\linewidth}{!}{
    \begin{tabular}{l|c|c}
        \hline
        ~ & Offset Towards Center & Orientation Vector \\
        \hline
        Direct Regression  &  0.197 & 0.806  \\
        Random Guess  &  0.228 & 0.801  \\
        \hline
    \end{tabular}}
    \end{center}
    \caption{\label{tab:intro}\textbf{Mean absolute error of point wise predicted offsets and orientations, evaluated on ScanNet.} We see that direct regression is only slightly better than random guess on offset predictions and worse on orientation predictions.}
\end{table}

To handle this problem, we disentangle direct offset towards object centers into the following three parts: Local Canonical Coordinates (LCC), box scales and box orientations. 
First, we estimate \textbf{Local Canonical Coordinates (LCC)} and box scales instead of regressing orientations. In LCC, all objects are consistently aligned and centered. In comparison to conventional orientation regression like voteNet, regression of LCC is generally easier because points belonging to the same part of an object are mapped to similar LCCs, no matter how the object rotates. Based on this, our model only needs to solve a task that is similar to part segmentation, for which we have seen great success in recent years~\cite{wu2019pointconv,qi2017pointnet,huang2018recurrent}.   
Experiments show that our LCC predictions are far more accurate than direct offset. Therefore, these canonical votes could be directly used without any post-processing step like clustering. Similar ideas on regressing local coordinates have also been explored by previous methods like NOCS~\cite{wang2019normalized}. However, it requires an additional Mask-RCNN to do instance segmentation, and then finds a closed-form solution of translation, rotation and scale. This closed-form solution only exists for a single object. If there are multiple instances, no global closed-form solution exists as far as we know. 

To solve this problem, we design a \textbf{canonical voting} algorithm to find possible object orientations and centers in Euclidean space. Object bounding boxes are proposed by looking at those locations with high votes. However, there will be some votes that accidentally accumulate as false positives. In order to eliminate them, it is crucial to project proposed object coordinates back into canonical space and compare them with LCC predictions. We call this step as \textbf{LCC checking with back projection}.

We evaluate our approach on two challenging large-scale 3D scan datasets: ScanNet~\cite{dai2017scannet} and SceneNN~\cite{hua2016scenenn}, and one smaller indoor RGB-D dataset: SUN RGB-D~\cite{song2015sun}.
Our approach achieves state-of-the-art performance on ScanNet and SceneNN, with an absolute advance of \textbf{9.6} and \textbf{5.1} mAP, respectively. It also shows superior performance on SUN RGB-D benchmark. In addition, our experiments show that LCC regression and canonical voting are more robust over direct offset and orientation regression on detecting occluded objects. 

To summarize, our contributions are:
\begin{itemize}
    \item Bypassing orientation regression difficulties through Local Canonical Coordinates and Canonical Voting.
    \item Devising a back projection validation module to eliminate false positives, achieving high average precision.
    \item State-of-the-art performance on three 3D bounding box detection benchmarks.
\end{itemize}

\section{Related Work}

\subsection{3D Object Detection}
There are many previous methods that are able to predict 3D bounding boxes of objects.  COG~\cite{ren2016three} proposes to detect object by capturing contextual relationships among categories and layout. PointPillars~\cite{lang2019pointpillars} utilizes PointNets~\cite{qi2017pointnet} to learn a representation of point clouds organized in vertical columns (pillars). PointRCNN~\cite{shi2019pointrcnn} leverages two stages to do bottom-up 3D proposal generation and further refine these proposals. Frustum PointNet~\cite{qi2018frustum} first detects 2D bounding boxes in images and then back projects them into 3D space in order to get 3D bounding boxes. VoteNet~\cite{qi2019deep} and ImVoteNet~\cite{qi2020imvotenet} both utilizes a clustering and voting scheme, though their direct offset predictions are not accurate and need to be refined with another MLP network. PointFusion~\cite{xu2018pointfusion} predicts multiple 3D box hypotheses and their confidences, using the input 3D points as spatial anchors. GSPN~\cite{yi2019gspn} generate proposals by reconstructing shapes from noisy observations in a scene. There are also a bunch of instance segmentation methods~\cite{jiang2020pointgroup,han2020occuseg,yang2019learning} that predict instances without oriented bounding boxes.

\subsection{Voting Methods}
Recently, there are several methods that combined deep learning and Hough Voting procedure, for various tasks.  VoteNet~\cite{qi2019deep} and ImVoteNet~\cite{qi2020imvotenet} use a PointNet++\cite{qi2017pointnet} backbone and generate votes for each seed point. These votes are then clustered in Euclidean space to form proposals, which are fed into another MLP network to give final detections. PVNet~\cite{peng2019pvnet} regresses pixel-wise unit vectors pointing to the predefined keypoints and solves a Perspective-n-Point (PnP) problem for pose estimation in RGB images. Zhao et al.~\cite{zhao2021deep} parameterizes lines with slopes and biases, and perform Hough transforms to translate deep representations into the parametric domain, in order to detect semantic lines in 2D images. In addition, there are several methods based on point-pair features~\cite{hinterstoisser2016going,drost2010model} to do a general Hough transform on object 6D poses. They need to sample a quadratic number of point pairs, making them extremely slow in large scenes.


\section{Method}

\begin{figure*}[ht]
    \centering
    \includegraphics[width=0.8\linewidth]{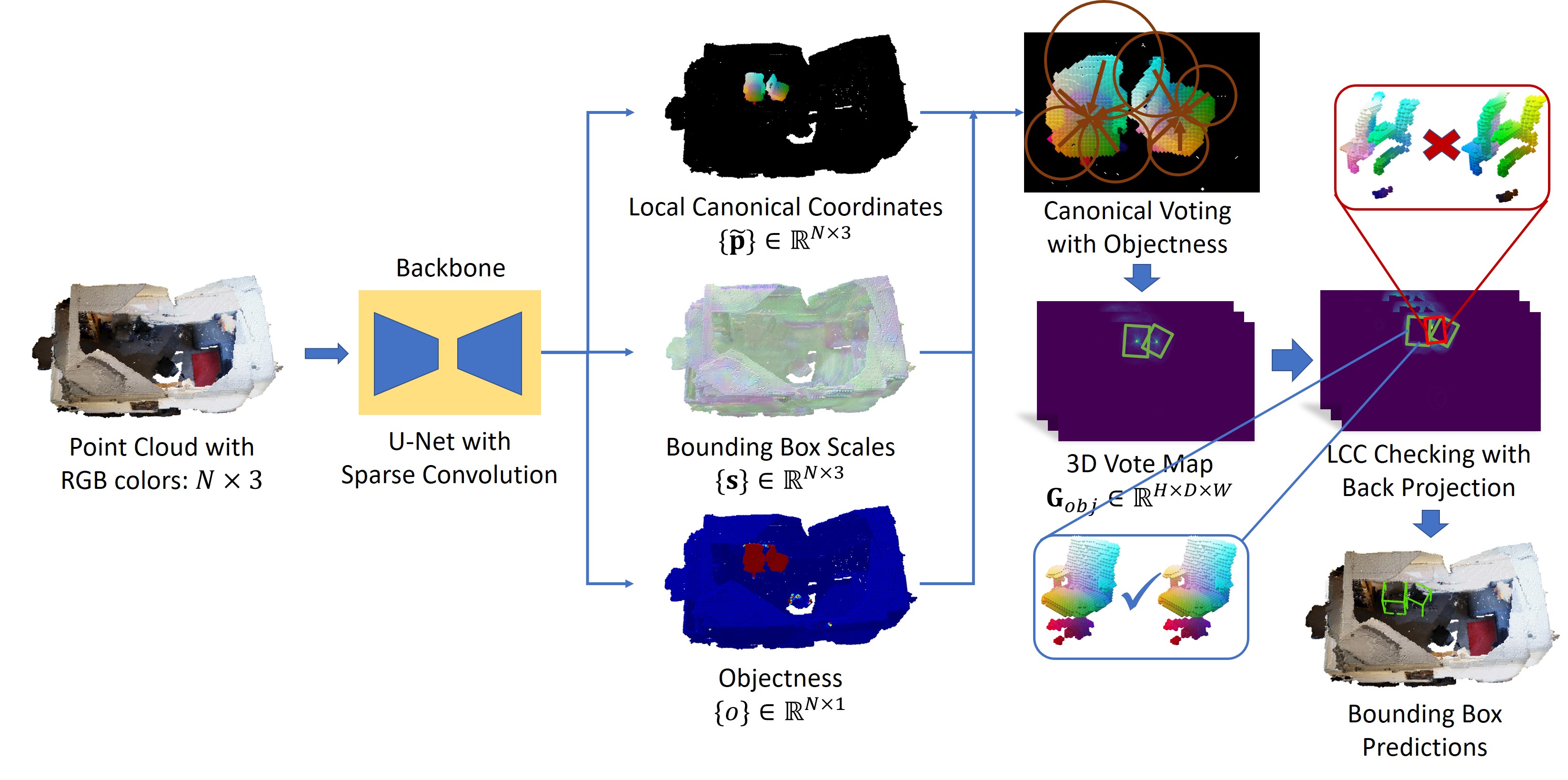}
    \caption{\textbf{Our model pipeline.} We first regress Local Canonical Coordinates (LCC), bounding box scales and objectness for each point; then an objectness weighted canonical voting algorithm is conducted to generate votes on 3D grids; finally, an LCC checking module with back projection validation is  leveraged  to  progressively generate bounding boxes from the 3D vote map, which is shown in bird's-eye view.}
    \label{fig:pipeline}
\end{figure*}
\subsection{Overview}
Figure~\ref{fig:pipeline} illustrates our detection pipeline. It can be split into three stages: first, we regress Local Canonical Coordinates, scales and objectness for each scene point, and we explain why it works in comparison to conventional direct offset regression~\cite{qi2019deep}; then a canonical voting algorithm is proposed to generate a vote map on 3D grids; finally,  an LCC back projection checking module is leveraged to progressively eliminate false positives and generate bounding boxes. To make it easier to understand, we will first describe the procedure for single-class object prediction and then discuss how it can be extended to multiple classes.

\subsection{Regressing Local Canonical Coordinates}
\begin{figure}[ht]
    \centering
    \includegraphics[width=0.8\linewidth]{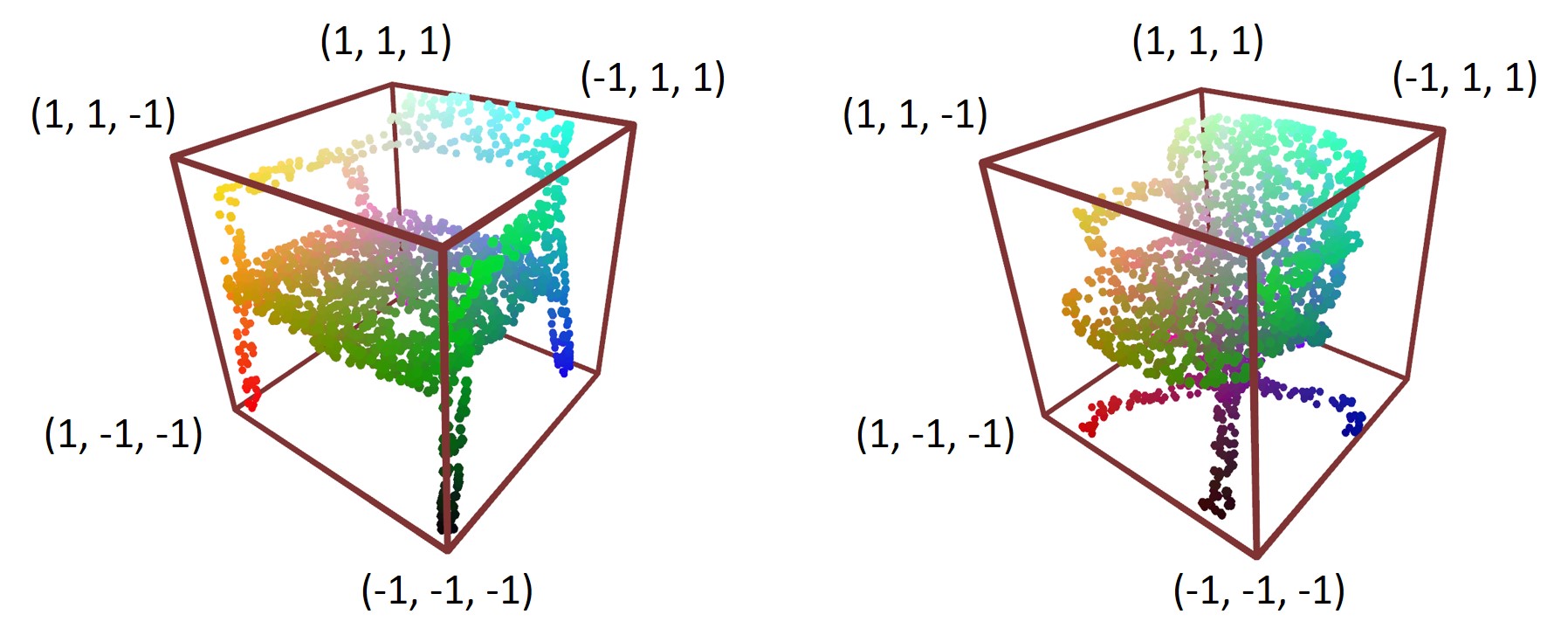}
    \caption{\textbf{The Local Canonical Coordinate (LCC) space.} 
    RGB colors indicate the $(x, y, z)$ positions in the LCC.}
    \label{fig:lcc}
\end{figure}

Inspired by NOCS~\cite{wang2019normalized}, we propose to regress Local Canonical Coordinates. Local Canonical Coordinates (LCC), similar to NOCS, is defined as a 3D space contained within a unit cube i.e., $\{x, y, z\} \in [-1, 1]$. In LCCs, all the models are consistently aligned and centered. Figure~\ref{fig:lcc} shows an example of LCCs for two chairs.

For each object bounding box, within which LCCs are linked to world coordinates by the bounding box parameters:
\begin{align}
\label{eq:lcc}
    \mathbf{p} &= \Psi_{\mathbf{s}, \alpha, \mathbf{t}}(\Tilde{\mathbf{p}})\notag\\
    &= \text{diag}(\mathbf{s})\cdot \text{R}_y(\alpha) \cdot\Tilde{\mathbf{p}} + \mathbf{t} \notag\\
    &= \begin{bmatrix}
    s_x & 0 & 0 \\ 
    0 & s_y & 0 \\
    0 & 0 & s_z \\
    \end{bmatrix} \cdot \begin{bmatrix}
    cos(\alpha) & 0 & -sin(\alpha) \\ 
    0 & 1 & 0 \\
    sin(\alpha) & 0 & cos(\alpha) \\
    \end{bmatrix} \cdot \Tilde{\mathbf{p}} + \begin{bmatrix}
    t_x \\ 
    t_y \\
    t_z \\
    \end{bmatrix},
\end{align}
where $\mathbf{p}\in \mathbb{R}^3$ is world coordinates and $\Tilde{\mathbf{p}}\in \mathbb{R}^3$  is Local Canonical Coordinates (LCC). $\mathbf{s} = [s_x, s_y, s_z]^T$ are bounding box scales, $\alpha$ is the heading angle of the bounding box around gravity axis, which is $y$-axis in our coordinate system. We follow previous works~\cite{qi2019deep} and only consider the heading angle around the gravity axis. Nonetheless, our model can be applied to full 3D rotations, showing an advancement on current 6D pose estimation benchmarks, with more details in our supplementary. $\mathbf{t} = [t_x, t_y, t_z]^T$ are bounding box centers. $\textrm{R}_y(\cdot)$ is an operator that generate a rotation matrix around $y$-axis given the heading angle.

Every point of an object in world coordinates can be transformed uniquely to LCC by Equation~\ref{eq:lcc}. In the following sections, we will denote Equation~\ref{eq:lcc} as $\mathbf{p} = \Psi_{\mathbf{s}, \alpha, \mathbf{t}}(\Tilde{\mathbf{p}})$.

Then, we regress LCCs and bounding box scales for each point on objects. Specifically, given a point cloud $\{\mathbf{p}_i\}_{i=1}^N$, we point-wise predict $\mathbf{s}_i, \Tilde{\mathbf{p}}_i$, with the following loss:

\begin{align}
\label{eq:loss}
    L_{reg} = &\sum_{j=1}^{|\{B_j\}|}\sum_{i=1}^N (\|\mathbf{s}_j^* - \mathbf{s}_i\| + \|\Psi_{\mathbf{s}_j^*, \alpha_j^*, \mathbf{t}_j^*}^{-1}(\mathbf{p}_i) - \Tilde{\mathbf{p}}_i\|) \notag
    \\& \cdot \mathbbm{1}(\mathbf{p}_i \text{ on object } j\text{'s surface}),
\end{align}
where ${\mathbf{s}_j^*, \alpha_j^*, \mathbf{t}_j^*}$ are ground truth scale, heading angle and translation parameters of bounding box $B_j$, respectively. $\mathbf{p}_i \text{ on object } j\text{'s surface}$ indicates whether the point $\mathbf{p}_i$ is on the surface of the object $j$ with bounding box $B_j$. 

It is worth mentioning that our LCC representation is invariant under rotations, while direct offsets are not. We do not predict rotations $\alpha_i$ and object centers $\mathbf{t}_i$, as these two parameters will be generated from the canonical voting stage described in the next section.

\begin{figure*}[ht]
    \centering
    \includegraphics[width=0.9\linewidth]{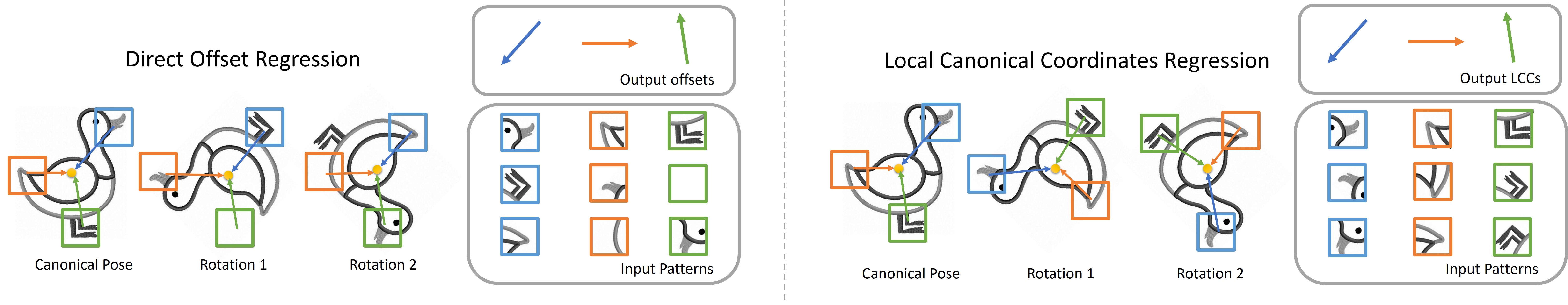}
    \caption{\textbf{An illustration on why LCC works better than direct offset regression.} \textbf{Left}: direct offset regression maps different parts of the object to the same output offsets. \textbf{Right}: LCC regression maps the same part to the same LCC in canonical pose, regardless of the rotation. Therefore, it is easier for a network to regress LCCs by doing a part segmentation task. Image from \href{http://clipart-library.com/}{Clipart Library}.}
    \label{fig:analy}
\end{figure*}

\paragraph{Why Local Canonical Coordinates?} At a first glance, it  seems that LCCs are indirect and hard to predict. However, this is not the case. Take a 2D image with rotations as an example as shown in Figure~\ref{fig:analy}, and suppose we would like to output direct offsets/LCCs for each pixel with 2D convolutional networks. In the left of Figure~\ref{fig:analy}, we see how direct offset regression works: each local pattern is mapped into its corresponding direction. When the image rotates, different parts of the duck are mapped to the same output offset (indicated by the same color), since direct offset does not ``go with'' rotation. This makes it tough to identify different offsets based on such divergent input patterns. 
In contrast, we can see how LCC regression works from the right of Figure~\ref{fig:analy}: no matter how the image rotates, patterns that belong to the same part (e.g. beak) are always mapped to the same LCC in canonical pose. This makes it easier to learn the relationship between inputs and outputs. The network only needs to classify different parts (with a few of rotated versions) of the object in order to output corresponding LCCs.

\paragraph{Object Symmetry} Our LCC representation would result in large errors for symmetric objects (e.g. tables, trash bins) as it is defined for a single orientation. To resolve this, we follow the common practice~\cite{wang2019normalized,hodan2020epos,wang2019densefusion} by using a variant of loss function. 
We calculate the loss $L_{reg}$ for all symmetric reflections of an object, and then take the \textit{minimum} of them.

\subsection{Canonical Voting with Objectness}
\label{sec:vote}
Next, we propose a canonical voting algorithm that produces a vote map  indicating the likelihood of the existence of any object. In order to filter out those votes from points that do not belong to any objects, an additional objectness score $o_i \in [0, 1]$ is predicted for each point, optimized by Cross Entropy loss with ground-truth objectness $o_i^*\in \{0,1\}$.
$o_i^* = 1$ if points are on any instance; $o_i^* = 0$ otherwise.

\begin{figure}[ht]
    \centering
    \includegraphics[width=0.7\linewidth]{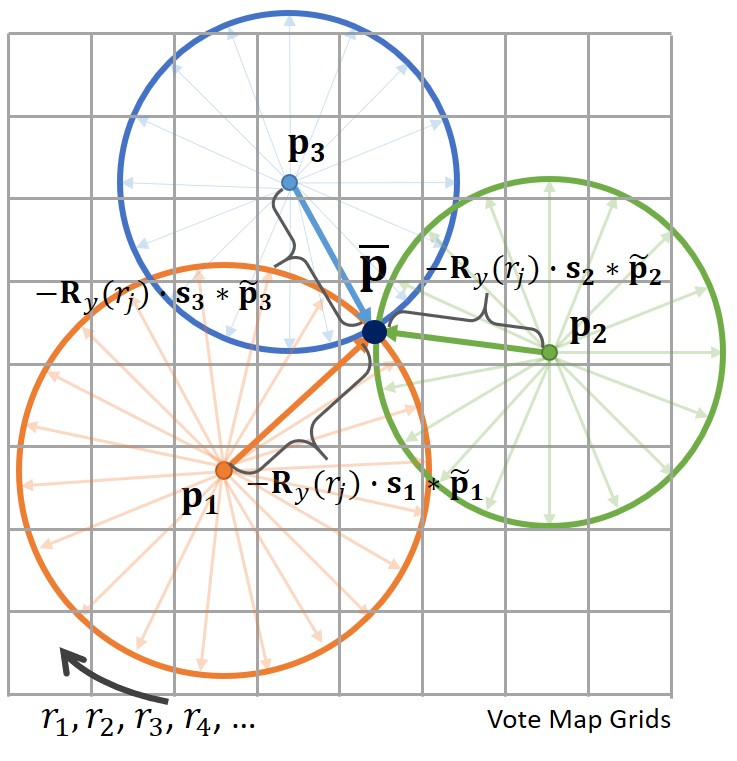}
    \caption{\textbf{Canonical Voting Process in Bird's-eye view. }  For each point, we estimate its LCC $\Tilde{\mathbf{p}}$ and bounding box scales $\mathbf{s}$. Then for each possible bounding box orientations $r_1,r_2,r_3,r_4,\dots$, we generate a vote towards the box center $\ \mathbf{p}$ by subtracting rotated $\mathbf{s} * \Tilde{\mathbf{p}}$ from its world coordinate $\mathbf{p}$. As last, these votes are accumulated in a predefined 3D grid by trilinear interpolation.}
    \label{fig:voting}
\end{figure}

Once we have $\mathbf{s}_i, \Tilde{\mathbf{p}}_i, o_i$, every point votes to its corresponding bounding box center, for every possible orientation. To accumulate votes, we discretize continuous Euclidean space into grids $\mathbf{G}_{obj}\in\mathbb{R}^{H\times D\times W}$, where $H,D,W$ is determined by a predefined grid interval $\tau$ and the extent of input point clouds. Two additional grids $\mathbf{G}_{rot}\in\mathbb{R}^{H\times D\times W}$, $\mathbf{G}_{scale}\in\mathbb{R}^{H\times D\times W \times 3}$ are leveraged to capture voted heading angles and box scales, respectively. Details are given in Algorithm~\ref{alg:vote} and this step is visualized in Figure~\ref{fig:voting}.

\begin{algorithm}[ht] 
\begin{algorithmic}[1] 
    \STATE \textbf{Input}: For each point $i=1,\dots,N$, scale $\mathbf{s}_i$, LCC $\Tilde{\mathbf{p}}_i$, objectness $o_i$, world coordinate $\mathbf{p}_i$.
    \STATE \textbf{Output}: Accumulated votes for objectness $\mathbf{G}_{obj}$, rotation $\mathbf{G}_{rot}$ and scale $\mathbf{G}_{scale}$.
    \FOR{$i = 1,\dots, N$}
        \STATE $\mathbf{v}_i = \mathbf{s}_i * \Tilde{\mathbf{p}}_i$, where $*$ is element-wise product.
        \FOR{$j = 1,\dots, K$}
            \STATE Find possible orientation $r_j = \frac{j}{K}*2\pi$.
            \STATE Find possible center $\Bar{\mathbf{p}}_i = \mathbf{p}_i - \textrm{R}_y(r_j)\cdot \mathbf{v}_i$.
            \STATE Find $\Bar{\mathbf{p}}_i$'s $2^3 = 8$ discrete grid neighborhoods $\mathcal{N}$.
            \STATE Add $o_i$ to $\mathcal{N}$ on $\mathbf{G}_{obj}$ with trilinear interpolation.
            \STATE Add $o_i\cdot r_j$ to $\mathcal{N}$ on $\mathbf{G}_{rot}$ with trilinear interpolation.
            \vspace{-0.06\linewidth}
            \STATE Add $o_i\cdot \mathbf{s_i}$ to $\mathcal{N}$ on $\mathbf{G}_{scale}$ with trilinear interpolation.
        \ENDFOR
    \ENDFOR
    \STATE Normalize rotation and scale by element-wise division: $\mathbf{G}_{rot} = \frac{\mathbf{G}_{rot}}{\mathbf{G}_{obj}}, \mathbf{G}_{scale} = \frac{\mathbf{G}_{scale}}{\mathbf{G}_{obj}}$.
    \caption{Canonical voting process.}
    \label{alg:vote} 
\end{algorithmic}
\end{algorithm}

In line 4 of Algorithm~\ref{alg:vote}, we generate possible point offset to object center without considering rotations. Since every object is considered to be rotated along gravity direction, we rotate the offset for every possible rotation in line 6 up to a predefined resolution K. In line 7, possible object centers are found by inverting Equation~\ref{eq:lcc}. Then a vote will be accumulated in a predefined grid through trilinear interpolation. Finally, the voted heading angle map and scale map are weighted by objectness in line 14. 120 possible rotations are tested for each point, and the algorithm is highly paralleled and implemented on GPU. This step is linear in the number of points, and it generally runs within 30ms.

\paragraph{Every Point is a First-class Citizen}
We can see that every single point would participate in the canonical voting process. As a consequence, our model is more robust than previous methods on detecting partial or occluded objects, as each point would generate a candidate vote. In contrast, previous two-stage methods like VoteNet generate proposals by clustering or sub-sampling, where occluded objects are more likely to be missed. This will be investigated in detail in our experiments.

\subsection{LCC Checking with Back Projection for Bounding Box Generation}

Next, we identify the peaks in the vote map $\mathbf{G}_{obj}$ and generate bounding boxes. Importantly, false positives are eliminated by operating LCC checking with back projection.

\begin{figure}[ht]
    \centering
    \includegraphics[width=0.8\linewidth]{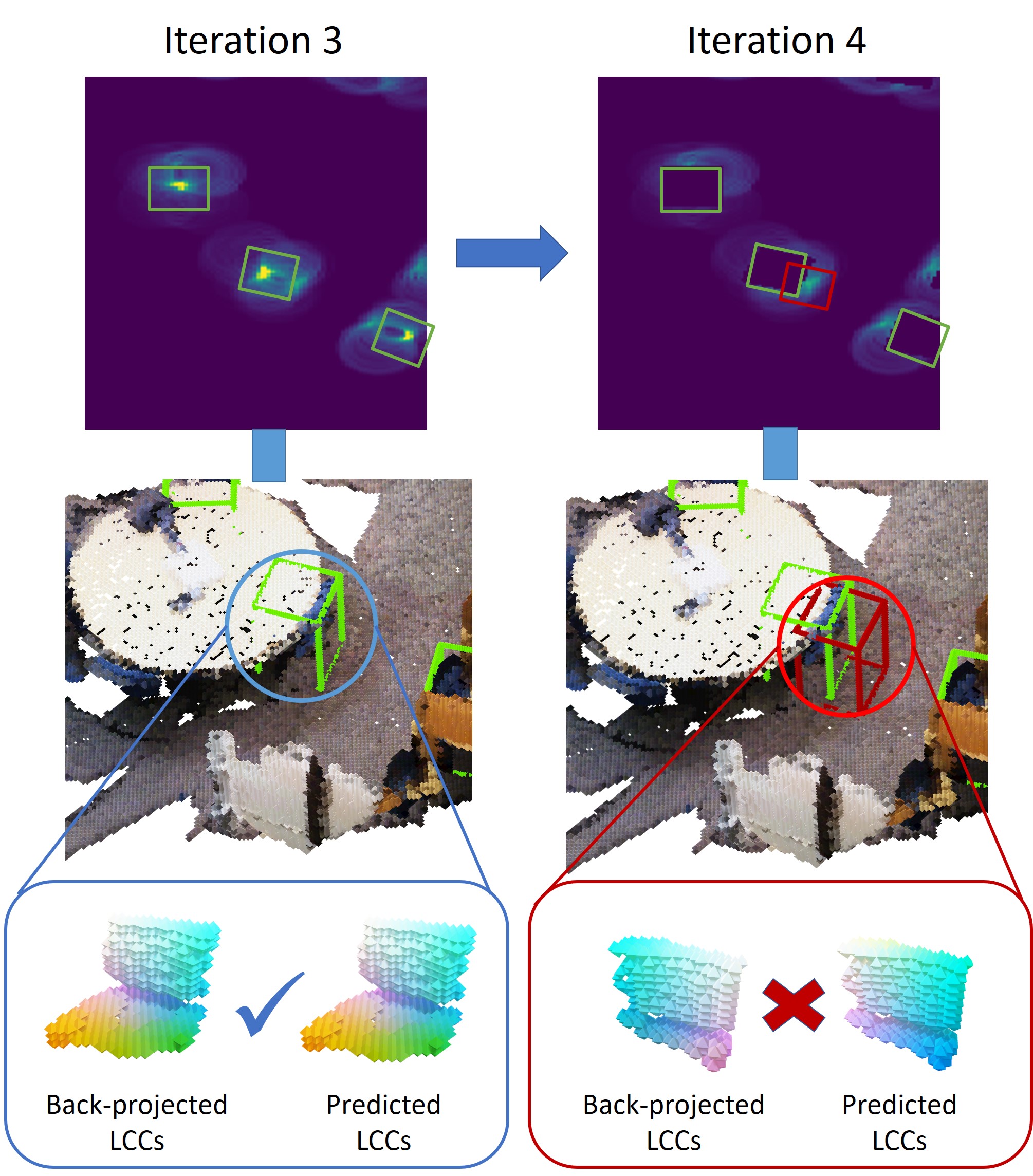}
    \caption{\textbf{LCC Checking with Back Projection Module.} Once the 3D vote map is proposed, we greedily find the peaks and generate candidate bounding boxes. LCC coordinates are visualized as RGB colors. At iteration 3, three true positive (in green) boxes for (partial) chairs are generated. When their coordinates are back projected to LCCs, they are consistent with network predictions. At iteration 4, due to the exhaustive orientation search in the canonical voting process, there is a false positive box (in red). It is eliminated by comparing the LCCs back projected with the candidate bounding box and the LCCs from network predictions.  }
    \label{fig:generate}
\end{figure}

\paragraph{LCC Checking with Back Projection}
Because of the exhaustive orientation search in canonical voting process, there will be some false peaks in the resulting vote map $\mathbf{G}_{obj}$, as shown in Figure~\ref{fig:generate}. In order to eliminate them, we leverage an LCC checking process with back projection. Specifically, we first generate a bounding box candidate according to the peak of $\mathbf{G}_{obj}$, $\mathbf{G}_{rot}$ and $\mathbf{G}_{scale}$.  Then, we back project original points into Local Canonical Coordinates $\Tilde{\mathbf{p}}'$ according to the current candidate bounding box. Points outside the candidate bounding box are discarded. Finally, we check whether the projected LCCs $\Tilde{\mathbf{p}}'$ are consistent with dense LCC predictions $\Tilde{\mathbf{p}}$ of the network. This step is effective and crucial and most false positives are eliminated in this step, reaching a high precision. If the candidate box is placed wrongly, then its orientation will be inaccurate and the projected LCCs will be inconsistent with those of network predictions. The full algorithm is demonstrated in Algorithm~\ref{alg:bbox}.

\begin{algorithm}[ht] 
\begin{algorithmic}[1] 

    \STATE \textbf{Input}: Accumulated votes for objectness $\mathbf{G}_{obj}$, rotation $\mathbf{G}_{rot}$ and scale $\mathbf{G}_{scale}$; for $i=1,\dots,N$, scale $\mathbf{s}_i$, LCC $\Tilde{\mathbf{p}}_i$, objectness $o_i$, point $\mathbf{p}_i$.
    \STATE \textbf{Output}: Predicted bounding box set $\mathbf{B}$.
    \STATE Initialize $\mathbf{B} = \{\}$.
    \WHILE{True}
        \STATE $[h,d,w]^T=\mathrm{argmax}(\mathbf{G}_{obj})$.
        \IF{$\mathbf{G}_{obj}[h,d,w] < \delta$}
            \STATE \textbf{break}
        \ENDIF
        \STATE Convert grid index $[h,d,w]^T$ to world coordinate $\mathbf{t}$.
        \STATE $\mathbf{s}:=\mathbf{G}_{scale}[h,d,w]$, $\alpha:=\mathbf{G}_{rot}[h,d,w]$.
        \STATE $err := 0$, $pos := 0$, $cnt := 0$, $o_{sum} := 0$.
        \FOR{$i = 1,\dots, N$}
            \STATE Let $\Tilde{\mathbf{p}}_i' = \Psi_{\mathbf{s}, \alpha, \mathbf{t}}^{-1}(\mathbf{p}_i)$.
            \IF{$-1 < \Tilde{\mathbf{p}}_i' < 1$}
                \STATE $cnt := cnt + 1$.
                \STATE Convert world coordinate $\mathbf{p}_i$ to grid index $[h_i,d_i,w_i]^T$.
                \STATE $\mathbf{G}_{obj}[h_i,d_i,w_i] := 0$.
                \IF{$o_i > 0.3$}
                    \STATE $pos := pos + 1$.
                    \STATE $o_{sum} := o_{sum} + o_i$.
                    \STATE $err := err + o_i\cdot \|\Tilde{\mathbf{p}}_i - \Tilde{\mathbf{p}}_i'\|_2$.
                \ENDIF
            \ENDIF
        \ENDFOR
        \IF{$pos > \beta\cdot cnt$ and $\frac{err}{o_{sum}} < \gamma$}
            \STATE Add bounding box $\{\mathbf{s}, \alpha, \mathbf{t}\}$ to $\mathbf{B}$.
        \ENDIF
    \ENDWHILE
    \caption{Bounding Box Generation with LCC Back Projection Checking.}
    \label{alg:bbox} 
\end{algorithmic}
\end{algorithm}
In Algorithm~\ref{alg:bbox}, from line 4 to line 6, a bounding box center is proposed greedily from the peaks in $\mathbf{G}_{obj}$, which is repeated until the vote count is below some threshold $\delta$. Then we read the scale and heading angle at the corresponding location of $\mathbf{G}_{scale}$ and $\mathbf{G}_{rot}$, and the world coordinate of bounding box center is retrieved by converting discrete grid index to continuous coordinate. 

Next, we back project all the points back into LCCs $\Tilde{\mathbf{p}}'$ according to the current candidate bounding box in line 13. If $\Tilde{\mathbf{p}}'$ is within the proposed bounding box, we set its vote count to 0 in $\mathbf{G}_{obj}$, so that it will never be detected again. Afterwards, two additional validation steps are leveraged to filter out those false positives. The first is to check if there are enough positive points (with probability $\beta$) within the bounding box, and the other is to check whether the back projected LCCs $\Tilde{\mathbf{p}}'$ are consistent with network LCC predictions $\Tilde{\mathbf{p}}$ in line 21. This process is visualized in Figure~\ref{fig:generate}.

\subsection{Extension to Multiple Classes}
Extension to multiple classes is quite straight-forward. In addition to scale $\mathbf{s}_i$, LCC $\Tilde{\mathbf{p}}_i$ and objectness $o_i$, a per-point class score $c_i$ is predicted, trained with cross entropy loss. The bounding box generation algorithm is almost identical to Algorithm~\ref{alg:bbox}, except that we determine the class with the majority votes of points within the bounding box.


\section{Experiments}
In this section, we first evaluate on three object detection benchmarks: ScanNet~\cite{dai2017scannet}, SceneNN~\cite{hua2016scenenn} and SUN RGB-D~\cite{song2015sun} dataset. Then, we conduct detailed analysis to verify our algorithm designs.

\subsection{Experimental Setup}

\paragraph{Dataset} 
ScanNet is a richly annotated dataset of 3D reconstructed meshes of indoor scenes. Since original ScanNet does not provide amodal or oriented bounding box annotation, we use the oriented bounding box labels provided by Scan2CAD~\cite{Avetisyan_2019_CVPR}, where they annotate 14K+ oriented object bounding boxes for all 1506 scenes in ScanNet. Scan2CAD contains 9 common categories with rotation-aligned models. 

SceneNN~\cite{hua2016scenenn} is an RGB-D scene dataset consisting
of 100 scenes. 
We follow~\cite{hua2018pointwise} to evaluate on 76 scenes with orientated bounding box annotations. We directly transfer the model trained on ScanNet in order to see its generalization ability. Categories that are common to both SceneNN and ScanNet are reported.

SUN RGB-D~\cite{song2015sun} is a single-view RGB-D dataset
for scene understanding. It consists of ~5K RGB-D training images annotated with oriented 3D bounding boxes for 37 object categories. 
We follow a standard evaluation protocol and report performance on the 10 most common categories.

\paragraph{Compared Methods} We compare with a number of prior methods on 3D bounding box estimation: PointFusion~\cite{xu2018pointfusion}, GSPN~\cite{yi2019gspn}, F-PointNet~\cite{qi2018frustum}, VoteNet~\cite{qi2019deep}, H3DNet~\cite{zhang2020h3dnet} and BRNet~\cite{cheng2021back}.
Note that both PointFusion and F-PointNet require additional 2D corresponding images, which are not annotated by~\cite{hua2018pointwise} on SceneNN. Therefore, only GSPN, VoteNet and H3DNet results are reported on SceneNN. DSS~\cite{song2016deep} and COG~\cite{ren2016three} are also reported on SUN RGB-D.

\paragraph{Evaluation Metrics}  Evaluation metric is average precision with 3D bounding box IoU under various thresholds. We also find that our network performs better when trained separately for each category, due to the similarity of LCCs within each category. Therefore, we report results under two settings: separate training for each category and joint training for all categories.

\paragraph{Implementation Details}
Our detection pipeline takes an RGB point cloud as input. To augment data, we follow the same strategy as \cite{Avetisyan_2019_CVPR} so that point clouds are rotated around gravity direction four times ($90^{\circ}$ increments with $20^{\circ}$ random jitter). We follow the common practice to discretize point clouds into voxels with grid size 0.03 and adopt 3D Minkowski convolution networks with a similar structure as ResNet34.  The output size is 7NC + 1, consisting 3NC scales, 3NC LCC coordinates, NC + 1 class scores including backgrounds. We train the network from scratch with Adam optimizer, batch size 3 with an initial learning rate of 0.001. The hyperparameters in Algorithm~\ref{alg:bbox} are chosen with 5-fold cross-validation on training set: $\beta=0.2,\gamma=0.3,\delta=60$. We use heading angle directions $(cos(\alpha), sin(\alpha))$ rather than raw heading angle $\alpha$ during canonical voting and bounding box generation. Non-Maximum-Suppression with threshold 0.3 is leveraged on generated bounding boxes.

SUN RGB-D dataset does not provide segmentation labels but bounding boxes for each individual object. In order to train the objectness score, we consider all the points in a slightly enlarged (i.e., 1.2$\times$) bounding box belong to this object. Furthermore, because of the limited data (i.e., single RGB-D frames compared to multi-view reconstructed scenes in ScanNet), the generated vote map is not accurate enough. To solve this problem, instead of a deterministic bounding box generation procedure, we sample multiple box candidates with probability proportional to the vote map, and then leverage a refinement module to further refine these bounding boxes. 
For more details of this architecture, we refer the reader to the supplementary.

\begin{table*}[ht]
    \begin{center}
    \resizebox{0.95\linewidth}{!}{
    \begin{tabular}{l|ccccccccc|c}
        \hline
        ~ & Trashbin & Bathtub & Bookshelf & Cabinet & Chair & Display & Sofa & Table \& Desk & Others & mAP$_{50}$ \\
        \hline
        PointFusion~\cite{xu2018pointfusion} & \textbf{4.5}/- & 6.2/- & 4.6/- & 5.3/- & 25.1/- & 0.6/- & 3.8/- & 3.1/- & 6.6/- & 6.6/-\\
        GSPN~\cite{yi2019gspn} & 0.9/- & 5.4/- & 0.0/- & 0.5/- & 16.8/- & 0.2/- & 14.2/- & 5.1/- & 0.1/- & 4.8/-\\
        F-PointNet~\cite{qi2018frustum} & 0.0/2.3 & 7.9/9.2 & 3.8/6.9 & 12.8/14.7 & 39.3/34.1 & 0.0/\textbf{1.2} & 16.9/15.6 & 9.9/11.7 & 4.4/\textbf{6.6} & 10.6/11.4\\
        VoteNet~\cite{qi2019deep} & 2.3/5.2 & 7.4/\textbf{9.8} & 0.0/0.1 & 1.9/4.4 & 52.1/71.0 & \textbf{1.4}/0.1 & 2.5/0.5 & 17.0/10.2 & 8.5/6.4 & 10.3/11.9 \\
        H3DNet~\cite{zhang2020h3dnet} & 2.4/1.5 & \textbf{10.5}/3.2 & 1.1/0.5 & 9.8/4.4 & 23.7/28.7 & 0.2/0.0 & 23.7/3.2 & \textbf{31.5}/5.8 & 6.0/3.6 & 12.1/5.7 \\
        BRNet~\cite{cheng2021back} & 0.9/0.1 & 1.4/0.0 & 0.2/0.0 & 1.8/0.1 & 48.5/10.3 & 0.0/0.0 & 1.5/0.2 & 10.5/3.4 & \textbf{10.1}/2.7 & 8.3/1.9\\
        \hline
        Ours & 0.5/\textbf{32.6} & 0.5/6.9 & \textbf{14.2}/\textbf{8.9} & \textbf{17.0}/\textbf{15.2} & \textbf{57.6}/\textbf{75.5} & 0.1/0.0 & \textbf{39.5}/\textbf{31.3} & 3.2/\textbf{23.9} & 6.4/0.3 & \textbf{15.4}/\textbf{21.7} \\
        \hline
    \end{tabular}}
    \end{center}
    \caption{\label{tab:scannet}\textbf{3D oriented instance bounding box detection results on ScanNet val set, with Scan2CAD labels.} Results under both joint and separate training settings are reported, separated by slashes.}
\end{table*}

\begin{table}[ht]
    \begin{center}
    \resizebox{\linewidth}{!}{
    \begin{tabular}{l|ccccc|c}
        \hline
        ~ & Table \& Desk & Chair & Cabinet & Sofa & Display & mAP$_{50}$ \\
        \hline
        GSPN~\cite{yi2019gspn} & 8.4/- & 34.6/- & 0.2/- & 9.8/- & 0.0/- & 10.6/-\\
        VoteNet~\cite{qi2019deep} & \textbf{24.8}/27.3 & 65.2/78.0 & 1.5/1.0 & 15.4/2.4 & \textbf{0.7}/0.0 & \textbf{21.5}/21.7 \\
        H3DNet~\cite{zhang2020h3dnet} & 12.1/7.4 & 29.4/38.2 & 0.8/0.7 & 10.5/\textbf{14.8} & 0.0/0.0 & 10.6/12.2 \\
        BRNet~\cite{cheng2021back} & 17.3/5.2 & 58.3/13.3 & 0.4/0.0 & 0.8/0.1 & 0.0/0.0 & 15.4/3.7 \\
        \hline
        Ours  & 11.9/\textbf{43.4} & \textbf{68.1}/\textbf{79.6} & \textbf{2.5}/\textbf{2.2} & \textbf{22.8}/8.8 & 0.0/0.0 &  21.1/\textbf{26.8}  \\
        \hline
    \end{tabular}}
    \end{center}
    \caption{\label{tab:scenenn}\textbf{3D oriented instance bounding box detection results on SceneNN.} Our model achieves the highest mAP$_{50}$.}
\end{table}

\subsection{Evaluation on ScanNet and SceneNN}

We first report performance of our model on ScanNet validation set under two training settings: separately trained for each category and trained jointly for all categories. The results are listed in Table~\ref{tab:scannet}.
Our model outperforms previous methods under both training settings on Scan2CAD benchmark. It has an improvement of \textbf{3.3} mAP when trained jointly and \textbf{9.8} mAP when trained separately for each category. Our model outperforms baseline methods on Trashbin by \textbf{28.1} AP, Sofa by \textbf{15.8} AP, which is a large margin. Our model achieves even better performance when trained separately because it could benefit from Local Canonical Coordinate definitions that are similar and consistent in each category (Figure~\ref{fig:lcc}). In contrast, previous methods do not gain much or even perform worse when training separately.

\begin{figure}[ht]
    \centering
    \includegraphics[width=0.9\linewidth]{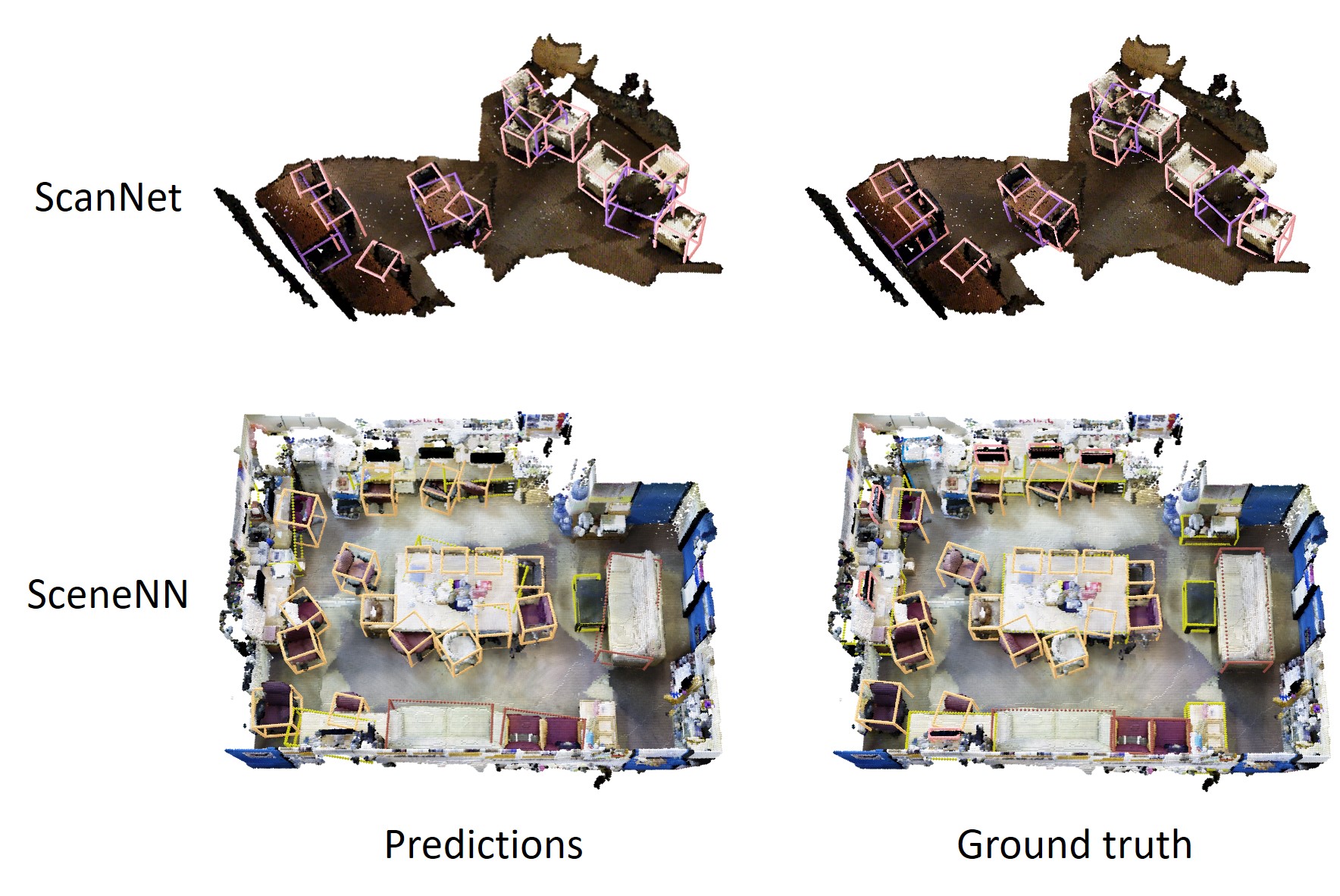}
    \caption{\textbf{Qualitative results on ScanNet and SceneNN.}}
    \label{fig:qual}
\end{figure}

Then we report our model's performance by directly evaluating on SceneNN dataset with network trained on ScanNet. The results are listed in Table~\ref{tab:scenenn}. Our method achieves highest mAP with good generalization ability. Some qualitative results in cluttered scenes are demonstrated in Figure~\ref{fig:qual}. 



\begin{figure}[ht]
    \centering
    \includegraphics[width=0.8\linewidth]{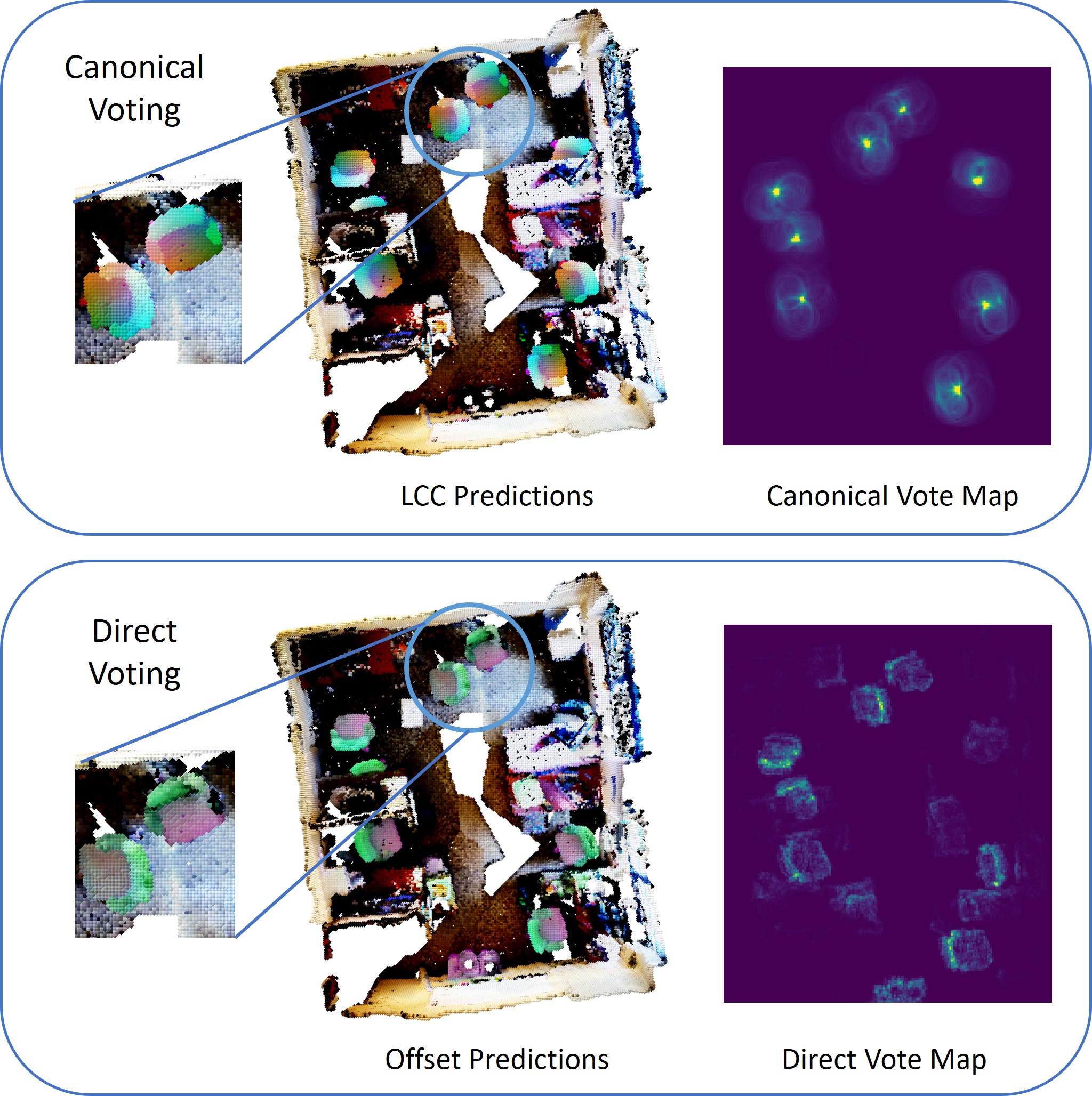}
    \caption{\textbf{LCC regression accumulates votes much better than direct offset regression.} LCC/offset predictions are visualized as RGB color tuples. In the top, we see that with LCC predictions, object centers are aggregated with good locality. On the contrary, direct offsets are not distinguishable on $xz$-plane (perpendicular to gravity axis) and object centers fail to accumulate with voting. }
    \label{fig:ablation}
\end{figure}

\subsection{Evaluation on SUN RGB-D}
Next, we evaluate on SUN RGB-D dataset, which contains single RGB-D frames instead of complete 3D scans. Quantitative results are listed in Table~\ref{tab:sunrgbd}, where our method also achieves state-of-the-art performance. All the compared methods are trained and evaluated jointly. 
With the help of our proposed canonical voting module, we are able to get higher mAP$_{50}$. This demonstrates that the voting center sampled from our vote map is more accurate than that of a simple Farthest Point Sampling (FPS) module.

\begin{table}[ht]
    \begin{center}
    \resizebox{0.5\linewidth}{!}{
    \begin{tabular}{l|c|c}
        \hline
        ~ & mAP$_{25}$ & mAP$_{50}$ \\
        \hline
        DSS~\cite{song2016deep} & 42.1 & -\\
        COG~\cite{ren2016three} & 47.6 & -\\
        F-PointNet~\cite{qi2018frustum} & 54.0 & - \\
        VoteNet~\cite{qi2019deep}  &  57.7 & 32.9 \\
        H3DNet~\cite{zhang2020h3dnet}  &  60.1 & 39.0  \\
        BRNet~\cite{cheng2021back} & 61.1 & 43.7 \\
        \hline
        Ours & \textbf{61.3} & \textbf{44.3} \\
        \hline 
    \end{tabular}}
    \end{center}
    \caption{\label{tab:sunrgbd}\textbf{Quantitative comparison on SUN RGB-D.}}
\end{table}
\subsection{Detailed Analysis}
\begin{figure}[ht]
    \centering
    \includegraphics[width=0.75\linewidth]{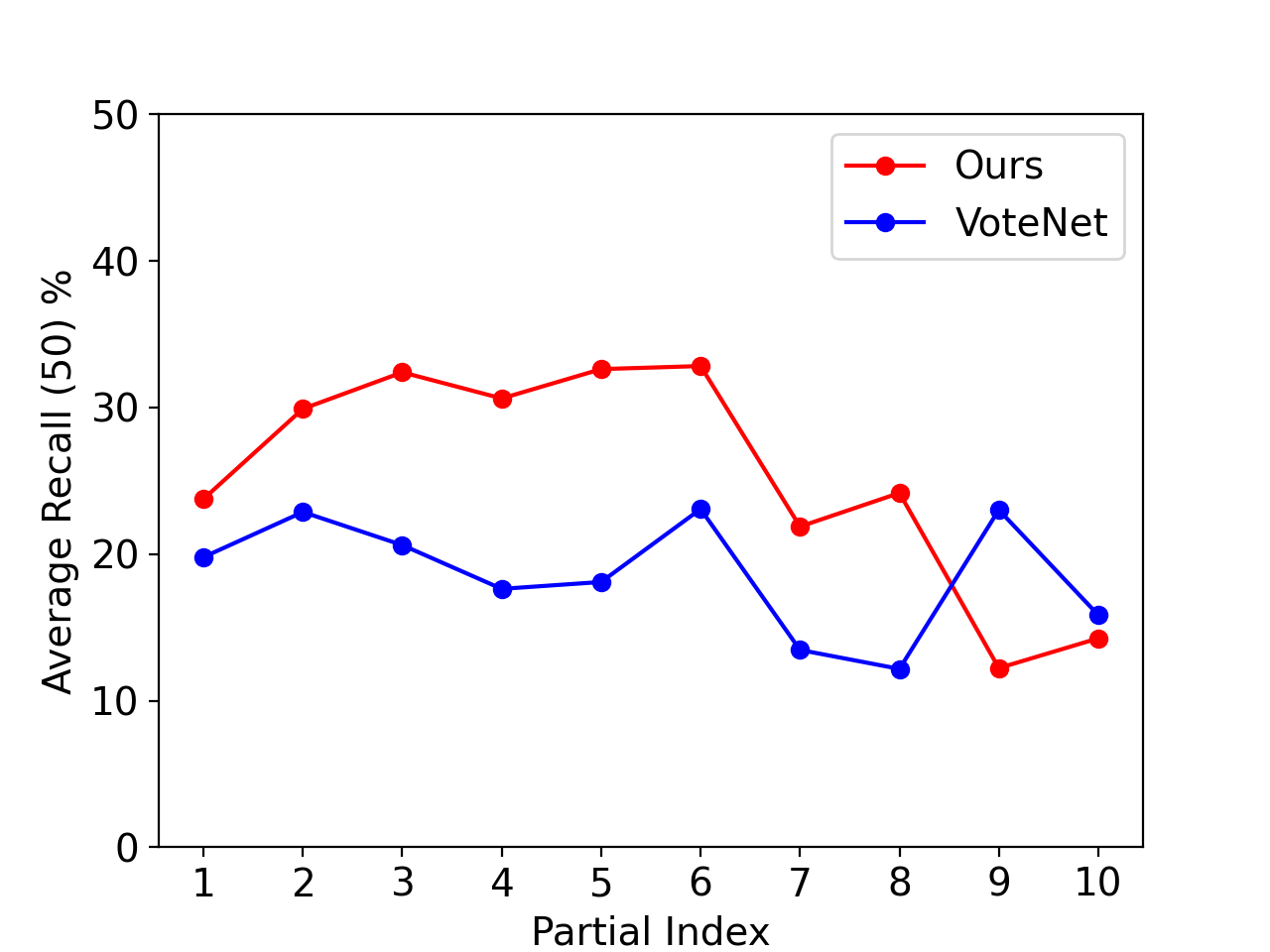}
    \caption{\textbf{Average Recall$_{50}$ comparison for partially occluded objects.} Our method achieves better performance on those partial/occluded objects.}
    \label{fig:robust}
\end{figure}

\paragraph{Robustness on Occluded/Partial Objects}
As we explained in Section~\ref{sec:vote}, each individual point is a first-class citizen, which has an advantage over previous methods on detecting occluded or partial objects, as these objects usually contain fewer points and any clustering or sub-sampling step would miss them by chance. To quantitatively illustrate this, we define a \textit{partial index} to be the number of points on an object, normalized by its bounding box volume. Intuitively, when an object has a small partial index, it is more likely to be occluded as fewer points are visible. We quantize these partial indexes into integers from 1 to 10, and plot the average recall for each discrete index, shown in Figure~\ref{fig:robust}. Our method gives higher AR$_{50}$ on severely occluded objects.
\paragraph{LCC + Canonical Voting vs. Direct Offset + Voting}
 We compare with our method with a baseline that implements the direct offset prediction and naive voting algorithm. Qualitative results are demonstrated in Figure~\ref{fig:ablation}. 
 Our model predicts LCCs accurately, and these LCCs can be directly utilized to vote in Euclidean space. On the contrary, direct offset predictions are inaccurate and fail to accumulate in Euclidean space. 
 Quantitative results are listed in Table~\ref{tab:ablation}.



\paragraph{How Does Back Projection Validation Help} 


In Table~\ref{tab:ablation}, we show the effectiveness of our LCC checking module with back projection. Without back projection, mAP experiences a large drop on both ScanNet and SceneNN datasets. 

\paragraph{Effect of Objectness} Objectness filters out many unlikely votes at the first stage. Here, we compare with a baseline algorithm where every point votes with weight one. Without objectness, mAP drops by 10.8 and 5.4 for ScanNet and SceneNN, respectively. 

\begin{table}[ht]
    \begin{center}
    \resizebox{0.9\linewidth}{!}{
    \begin{tabular}{l|c|c}
        \hline
        ~ & ScanNet & SceneNN \\
        \hline
        Ours (final) & \textbf{21.7} & \textbf{26.8}\\
        \hline
        Ours (direct voting)  &  0.0 & 0.0  \\
        Ours (no back projection check)  &  8.1 & 10.0  \\
        Ours (no objectness) & 12.2 & 21.4\\
        \hline
    \end{tabular}}
    \end{center}
    \caption{\label{tab:ablation}\textbf{Ablation study results, evaluated by mAP$_{50}$, with separate models.}}
\end{table}

\paragraph{Decoupled Orientation/Translation Error and Running Time}
The running time and decoupled
orientation/translation error of all compared methods are listed
in Table~\ref{tab:analy}. Our method outperforms previous
methods by a large margin in terms of orientation error.

\begin{table}[ht]
    \begin{center}
    \resizebox{\linewidth}{!}{
    \begin{tabular}{l|c|c|c|c}
        \toprule
        ~ & Ori. Error($^\circ$) & Trans. Error(m) & Proc. Time(s) & mAP$_{50}$ \\
        \midrule
        PointFusion~\cite{xu2018pointfusion} & 77.3 & 0.07 & \textbf{0.23} & 6.6 \\
        F-PointNet~\cite{qi2018frustum} & 62.5 & 0.09 & 0.46 & 10.6 \\
        VoteNet~\cite{qi2019deep} & 76.2 & 0.05 & 0.43 & 10.3 \\
        H3DNet~\cite{zhang2020h3dnet} & 65.4 & \textbf{0.03} & 0.78 & 12.1 \\
        BRNet~\cite{cheng2021back} & 79.1 & 0.05 & 0.47 & 8.3 \\
        \midrule
        Ours & \textbf{10.8} & 0.05 & 0.32 & \textbf{15.4}\\
        \bottomrule
    \end{tabular}}
    \end{center}
    \caption{\label{tab:analy}\textbf{Decoupled rotation/translation error (for mAP$_{50}$) and running time analysis on ScanNet with joint models.} Running time is reported on i9-7900X CPU with 1080Ti GPU.}
\end{table}



\section{Conclusion}
In this work, we propose a new method for robust oriented bounding box detection in large scale 3D scenes. We regress Local Canonical Coordinates (LCC) instead of direct offsets. Bounding box orientations are generated by Canonical Voting. LCC checking with back projection is leveraged to eliminate false positives. Results show that our model achieves state-of-the-art performance.

\section{Acknowledgements}
This work was supported by the National Key Research and Development Project of China (No. 2021ZD0110700), the National Natural Science Foundation of China under Grant 51975350, Shanghai Municipal Science and Technology Major Project (2021SHZDZX0102), Shanghai Qi Zhi Institute, and SHEITC (2018-RGZN-02046). This work was also supported by the  Shanghai AI development project (2020-RGZN-02006) and ``cross research fund for translational medicine'' of Shanghai Jiao Tong University (zh2018qnb17, zh2018qna37, YG2022ZD018).
{\small
\bibliographystyle{ieee_fullname}
\bibliography{egbib}
}

\appendix
\addcontentsline{toc}{section}{Appendices}
\include{supp}
\end{document}

%% file: supp.tex




\section*{Supplementary}



\section{More Ablation Analysis}
In this section, we conduct more ablation studies on multiple parameters that are used in Algorithm~\ref{alg:vote} and \ref{alg:bbox}.
\subsection{Visualization of Partial/Occluded Objects}
\begin{figure}[hb]
    \centering
    \includegraphics[width=\linewidth]{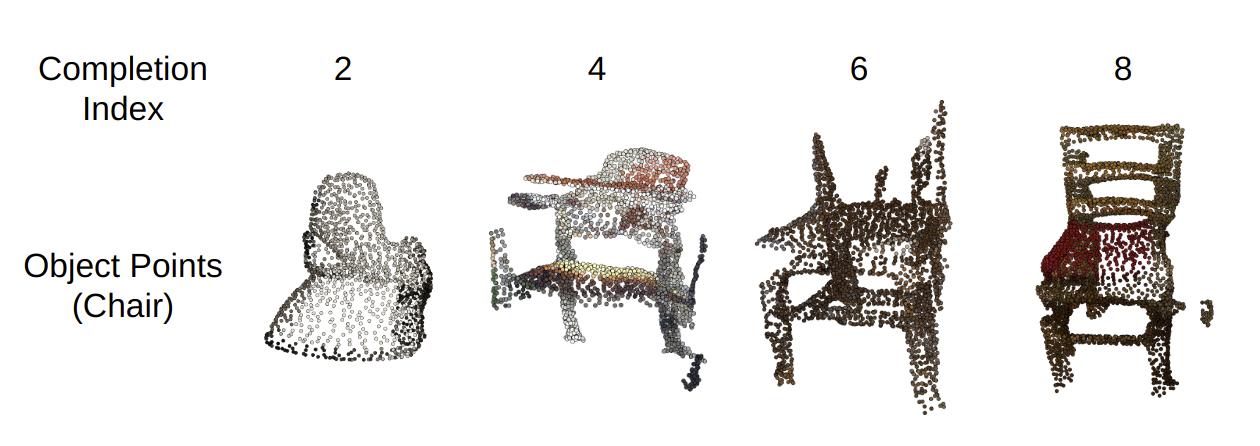}
    \caption{Point clouds of chairs with different partial indexes in ScanNet dataset. We see that small partial indexes refer to objects are occluded and partial, while objects with large partial indexes are more complete.}
    \label{fig:partial}
\end{figure}

As we mentioned in the main text, our method is robust in detecting partial/occluded objects. Here we visualize several objects of different partial indexes in Figure~\ref{fig:partial}. We can see that small partial indexes indicate occluded objects, while large partial indexes indicate complete objects. Our method gives higher average recall on partial objects compared with previous methods.

\subsection{Effects of $\delta$ in Bounding Box Generation }
In Algorithm~\ref{alg:bbox}, we iteratively generate bounding boxes whenever the maximum value of the heatmap is above a threshold $\delta$. Intuitively, with a small threshold, more object candidates are detected but may degrade the performance by introducing false positives at the same time. With a large threshold, our model is more confident on the detected bounding boxes, but may lower the recall by missing small objects.

\begin{figure}
    \centering
    \includegraphics[width=\linewidth]{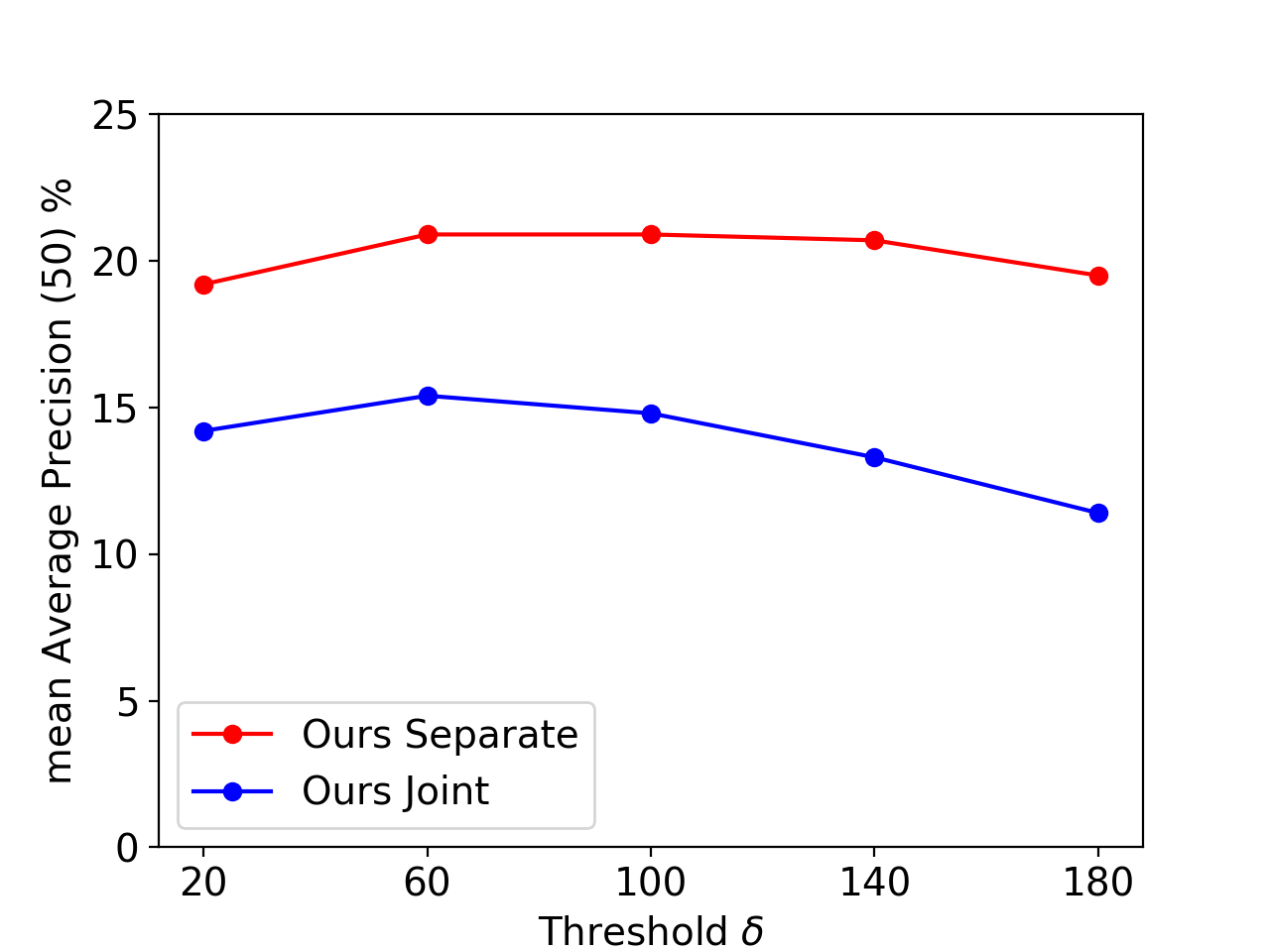}
    \caption{\textbf{mAP results on ScanNet
    val dataset under different $\delta$.}}
    \label{fig:delta}
\end{figure}

Here we investigate how this threshold influences the final mAP$_{50}$ on ScanNet val dataset. quantitative results are given in Figure~\ref{fig:delta}. 
The optimal threshold for $\delta$ lies around 60, with a balance between recall and precision.

\subsection{Effects of K in Canonical Voting Process}
\begin{figure}[h]
    \centering
    \includegraphics[width=0.9\linewidth]{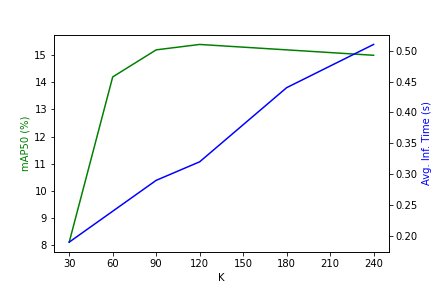}
    \caption{Average processing time and mAP$_{50}$ of our joint model on ScanNet. $K=120$ keeps a balance between time and accuracy.}
    \label{fig:ablationK}
\end{figure}
  
We report our joint model's average processing time per scan and mAP$_{50}$ for different values of $K$ of Algorithm~\ref{alg:vote} in Figure~\ref{fig:ablationK}. We see as $K$ increases, the processing time arises linearly because of the exhaustive search; while the mAP gets saturated after $K=120$.

\section{Extension to Full 3D Rotations}
Though we mainly conduct experiments on common indoor scenes, which usually contains only 1D rotations around the gravity axis, our voting scheme can be extended to full 3D rotation and be applied to NOCS REAL275 dataset (a 6D pose estimation benchmark). Specifically, we now treat every pixel in a given RGB-D frame as a 3D point which generates center offsets over $\frac{4\pi}{K}$ solid angles (K is resolution) instead of $\frac{2\pi}{K}$ circle angles. After candidate centers have been proposed, we collect all the points that contribute to these centers within some radius tolerance. The LCCs of these points are then used with the Umeyama algorithm to solve 3D rotations in closed form. Interestingly, our voting scheme on NOCS REAL275 achieves \textbf{17.0}, \textbf{40.7} and \textbf{45.8} mAP for (5$^\circ$, 5 cm), (10$^\circ$, 5 cm) and (10$^\circ$, 10 cm) metrics respectively, beating NOCS~\cite{wang2019normalized} baseline by a large margin.

\section{Details of Vote Map Guided Refinement Module on SUN RGB-D dataset}
\begin{figure*}[ht!]
    \centering
    \includegraphics[width=\linewidth]{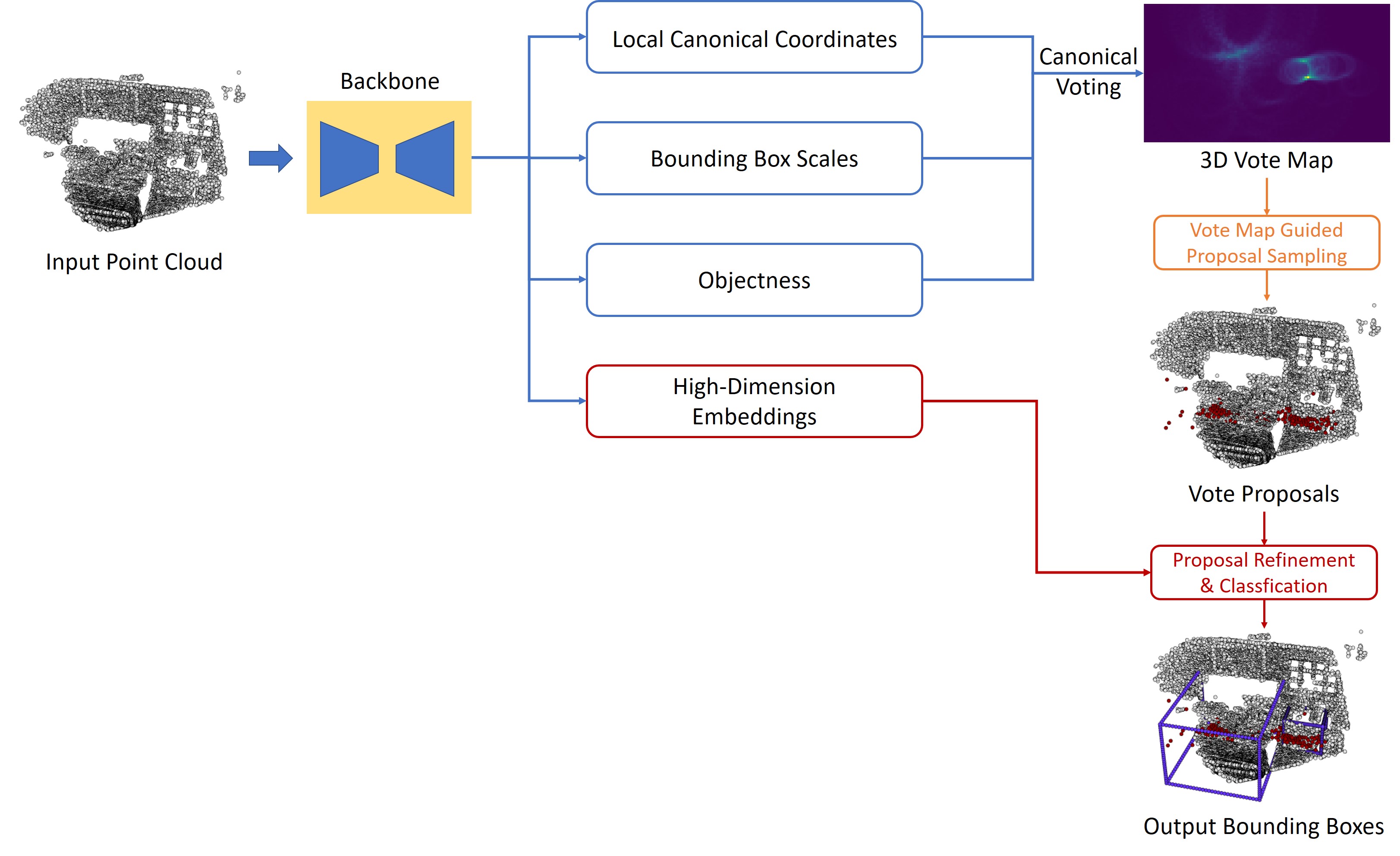}
    \caption{\textbf{The architecture of our method on SUN RGB-D dataset.}}
    \label{fig:pipe}
\end{figure*}

For SUN RGB-D dataset, due to the lack of symmetric information and limited training data, instead of a deterministic
bounding box generation procedure, we sample multiple box
candidates with probability proportional to the vote map,
and then leverage a refinement module to further refine these
bounding boxes. The pipeline is shown in Figure~\ref{fig:pipe}. The vote map generation is exactly the same as that in Algorithm~\ref{alg:vote}. Nevertheless, instead of direct bounding box generation with back-projection checking, we use a neural refinement module with learnable parameters to generate final bounding boxes. 
Specifically, we first generate 512 possible object center candidates according to the 3D vote map. The vote map is normalized to form a proper distribution. Then these vote proposals are then passed through a proposal refinement and classification module that is similar to BRNet. Every object proposal is aggregated with its surrounding back-traced representative points, using max pooling on their high dimensional embeddings generated by PointNet++. For more details of this refinement module, we refer the reader to BRNet~\cite{cheng2021back}.
